%% file: main.tex
\documentclass[11pt]{article}
\usepackage[margin=1in]{geometry}
\usepackage[T1]{fontenc}
\usepackage{times}
\usepackage{amsmath,amssymb,amsfonts,mathtools}
\usepackage{booktabs,multirow,makecell}
\usepackage{graphicx,subcaption}
\usepackage{xcolor}
\usepackage{placeins}
\usepackage{tikz}
\usetikzlibrary{arrows.meta,positioning,fit,backgrounds,shapes.geometric,calc}
\usepackage{pgfplots}
\pgfplotsset{compat=1.18}
\usepackage{algorithm}
\usepackage[noend]{algpseudocode}
\usepackage[numbers,sort&compress]{natbib}
\usepackage[colorlinks=true,citecolor=blue!60!black,linkcolor=blue!60!black,urlcolor=blue!60!black]{hyperref}
\usepackage[capitalise,noabbrev]{cleveref}


\makeatletter\@secpenalty=0\makeatother

\newcommand{\ours}{\textsc{SelfQuant}}

\newcommand{\Mz}{M_0}

\input{sections/numbers}

\title{SelfQuant: How Language Models Quantize Themselves Under a Fixed Memory Budget}
\author{Linghua Zhang \and Yunhe Li}
\hypersetup{pdftitle={SelfQuant: How Language Models Quantize Themselves Under a Fixed Memory Budget},pdfauthor={Linghua Zhang, Yunhe Li}}
\date{}

\begin{document}
\maketitle
\begin{abstract}
\input{sections/abstract}
\end{abstract}

\input{sections/introduction}
\input{sections/related_work}
\input{sections/method}
\input{sections/experiments}
\input{sections/results}

\input{sections/discussion}
\input{sections/conclusion}

{\small
\setlength{\bibsep}{2pt plus 0.3ex}
\bibliographystyle{plainnat}
\bibliography{refs}
}

\clearpage
\appendix
\input{sections/appendix}

\end{document}

%% file: sections/numbers.tex
\newcommand{\budgetGiB}{17.38}          %
\newcommand{\finalGiB}{17.38}            %
\newcommand{\finalBits}{3.88}           %

\newcommand{\klQz}{0.003263}            %
\newcommand{\klBelowQz}{10.9}           %
\newcommand{\klBestA}{0.003956}             %
\newcommand{\klBestB}{0.003269}             %
\newcommand{\klBestC}{0.003061}             %
\newcommand{\klBestD}{0.002921}             %
\newcommand{\klBestE}{0.002908}             %
\newcommand{\klIncA}{0.003956}              %
\newcommand{\klIncB}{0.003269}              %
\newcommand{\klIncC}{0.003061}              %
\newcommand{\klIncD}{0.002921}              %
\newcommand{\klIncE}{0.002908}              %
\newcommand{\validA}{8}              %
\newcommand{\validB}{6}
\newcommand{\validC}{8}
\newcommand{\validD}{5}
\newcommand{\validE}{7}
\newcommand{\klDropGen}{26.5}           %

\newcommand{\aimeMzA}{84.58}             %
\newcommand{\aimeMzB}{79.79}             %
\newcommand{\aimeMzAvg}{82.19}           %
\newcommand{\aimeSqA}{82.50}             %
\newcommand{\aimeSqB}{77.71}             %
\newcommand{\aimeSqAvg}{80.11}           %
\newcommand{\aimeRetain}{97.5}          %
\newcommand{\aimeGptqA}{81.25}         %
\newcommand{\aimeGptqB}{77.29}
\newcommand{\aimeGptqAvg}{79.27}
\newcommand{\aimeRtnA}{79.38}          %
\newcommand{\aimeRtnB}{69.58}
\newcommand{\aimeRtnAvg}{74.48}
\newcommand{\aimeAwqA}{81.67}          %
\newcommand{\aimeAwqB}{76.25}
\newcommand{\aimeAwqAvg}{78.96}
\newcommand{\sizeMz}{50.96}              %

%% file: sections/abstract.tex
Recent systems that design quantization for large language models rely on outside help, such as a frontier coding
agent. We ask whether a model can compress itself with no stronger model, API, teacher or gold label in the loop, and
study this as fixed-budget self-quantization. In \ours{}, the frozen BF16 model is the only intelligence: starting
from its own BF16 weights, eight independent sessions of the model per generation measure those weights with tools and
write programs that return mixed-precision configurations over 2, 3, 4, 8 and 16 bits, stored in no more bytes than
uniform 4-bit weight quantization. Each candidate is exported as a real low-bit checkpoint and scored by its KL
divergence from the BF16 model on 256 of the model's own reasoning traces; the lowest score below the incumbent's is
accepted. On Qwen3.8-27B, five generations lower this KL by $\klDropGen$\%, from $\klIncA$ to $\klIncE$, $\klBelowQz$\% below
uniform 4-bit GPTQ. At
$\finalBits$ bits per linear weight on average, the self-quantized model reaches $\aimeSqAvg$\% avg@16 on AIME24 and
AIME25, $\aimeRetain$\% of the BF16 model's $\aimeMzAvg$\% and above uniform 4-bit GPTQ ($\aimeGptqAvg$\%), AWQ-lite
($\aimeAwqAvg$\%) and RTN ($\aimeRtnAvg$\%) on both sets.

%% file: sections/introduction.tex
\section{Introduction}\label{sec:intro}

Weight-only post-training quantization is the standard way to fit large reasoning models into a fixed memory budget
\citep{gptq,awq}, and reasoning suffers most: accuracy falls as bits shrink and problems get harder
\citep{quanthurtsreasoning,reasoningcompress}, chains of thought grow without gaining accuracy \citep{lotfi} or loop
until the token budget runs out \citep{loopfix}, and perplexity can hide the damage \citep{llmkick,accnotall}.

The strongest recent designs come from outside help.
In QDA, a frontier coding agent (Claude Code with Claude Opus 5.5) composes a recipe from 21 PTQ methods, guided by
KL divergence to the BF16 model \citep{qda}.
Coding agents and LLMs also build compressed models \citep{compressionbench}, search quantization
group sizes \citep{autoqs}, or choose pruning settings \citep{selfpruner,autoprune,agentprune}.
In none of these systems is the decider verifiably the model being compressed.

We ask whether that help is needed: can a model quantize itself to a fixed memory budget with no stronger model, no
API, no teacher and no gold label in the loop?
The frozen BF16 model $\Mz$ is our system's only intelligence: it measures its own weights with tools, writes the
quantization configurations of its copies, and writes the reasoning traces on which those copies are judged.
The object improved is the model's \emph{deployable self at a fixed byte budget}, whose ceiling is $\Mz$'s own BF16
competence, not that of any external model.
The question is how much of that competence fits in the allowed bytes.

Self-quantization is a clean test of self-improvement without external help.
The target is internal and fixed, $\Mz$'s own next-token distribution, so an improvement cannot come from imitating a
stronger model, and its ceiling is known.
The constraint is byte-exact and checked by code, so a candidate cannot win by quietly using more memory.
And because every candidate quantizes the same original weights, the result of each generation is attributable to the
decisions the model wrote in that generation, informed by what it learned from earlier ones.

\ours{} (\cref{fig:overview}) runs in generations on Qwen3.8-27B, starting from the BF16 model itself.
A candidate assigns each module group 2, 3, 4, 8 or 16 bits, and its packed checkpoint may not be larger than uniform
4-bit GPTQ with group size 128.
In each generation, eight independent sessions of the frozen $\Mz$ read the incumbent's score and the archive of
earlier candidates, query tables of their own weights, and each write a program that returns one mixed-precision
configuration within the budget.
Every candidate is exported as a real low-bit checkpoint and scored by its KL divergence from $\Mz$ on 256 of $\Mz$'s
own reasoning traces, and the lowest score below the incumbent's is accepted.
Every candidate quantizes $\Mz$'s original weights, so errors do not compound.

Over five generations, the KL to $\Mz$ falls in every generation, by $\klDropGen$\% in total, and ends $\klBelowQz$\% below
uniform 4-bit GPTQ (\cref{tab:kl}).
On AIME24 and AIME25, which the loop never sees, the self-quantized model, at $\finalBits$ bits per linear weight on
average, reaches $\aimeSqAvg$\% avg@16, $\aimeRetain$\% of the BF16 model's $\aimeMzAvg$\%, and scores above uniform
4-bit GPTQ, AWQ-lite and RTN on both sets (\cref{tab:aime}).

Our contributions:
\begin{itemize}
  \item \textbf{A setting.} Self-improvement without external help as fixed-budget self-quantization by the frozen
  BF16 model (\cref{sec:method-setting}).
  \item \textbf{A method.} \ours{}: the frozen model writes byte-exact mixed-precision quantization programs for its
  own deployable copy, and a KL score on its own reasoning traces picks the incumbent
  (\cref{sec:method-loop,sec:method-gate}).
  \item \textbf{Results.} On Qwen3.8-27B, five self-written generations lower the KL to the BF16 model in every
  generation, and the resulting real low-bit model scores above uniform 4-bit RTN, GPTQ and AWQ-lite on AIME24 and
  AIME25 (\cref{sec:results}).
\end{itemize}

%% file: sections/related_work.tex
\section{Related Work}\label{sec:related}

\paragraph{LLM-designed compression.}
QDA~\citep{qda}, the closest work, has a frontier coding agent compose PTQ methods into fixed-format recipes scored by
KL divergence from the BF16 model.
CompressionBench~\citep{compressionbench} gives coding agents a storage limit and perplexity feedback,
AutoQS~\citep{autoqs} has an LLM propose group sizes that GPTQ executes, and TAP~\citep{tap} evolves mixed-precision proxies.
Self-Pruner~\citep{selfpruner}, AutoPrune~\citep{autoprune} and \citet{agentprune} use LLMs to guide pruning.
In none of these is the decider verifiably the model being compressed.

\paragraph{Mixed-precision allocation and post-training quantization.}
HAWQ~\citep{hawq} ranks layers by Hessian spectra, EvoPress~\citep{evopress} searches non-uniform profiles
evolutionarily, and AMQ~\citep{amq} pairs a quality predictor with memory-constrained search.
Quantizers include GPTQ~\citep{gptq}, AWQ~\citep{awq}, QuaRot~\citep{quarot} and GSQ~\citep{gsq};
self-calibration~\citep{selfcalib} calibrates on model-generated text.
We reuse GPTQ and RTN as back-ends and self-generated text as calibration data; our question is who
chooses the allocation, not how it is executed.

\paragraph{Quantization damage to reasoning.}
Perplexity can hide capability loss~\citep{llmkick}, and near-baseline accuracy can hide flipped answers~\citep{accnotall}.
Low bit-widths hurt reasoning models on math, science and code~\citep{quanthurtsreasoning,reasoningcompress},
lengthen chains of thought~\citep{lotfi} and cause loops and budget exhaustion~\citep{loopfix}.
Recovery methods distill the low-bit model on-policy from a full-precision teacher~\citep{opdlowbit,onq}; we train no weights.
We score candidates by their divergence from $\Mz$ on $\Mz$'s own long thinking-mode traces, not on generic text, and
report task accuracy on competition math.

\paragraph{Self-improvement and recursive self-improvement.}
The idea goes back to \citet{good1965} and the G\"odel machine~\citep{godelmachine}, which rewrites its own code
once a rewrite is proven useful.
Current systems replace proof with empirical validation and mostly change code around a fixed model:
STOP~\citep{stop} improves its own scaffold, the Darwin G\"odel Machine~\citep{dgm} edits its agent code with an archive,
and AlphaEvolve~\citep{alphaevolve} evolves programs under evaluator feedback.
Other work changes weights from the model's own outputs: self-rewarding models~\citep{selfrewarding},
SEAL~\citep{seal}, self-written edit templates~\citep{searchselfedit}, and self-written merging
algorithms~\citep{selfdeveloping}.
\ours{} takes the empirical acceptance test and archive from the first group and the object of change, its own weights,
from the second, without training or any outside model.
We claim no introspection, since LLMs fail to report their own quantization state~\citep{anosognosia}; the model
learns about its weights only through tools.
In short, \ours{} differs from prior LLM-designed compression in that the decider is verifiably the model whose own
weights are quantized, and candidates are judged only against that model's own next-token distributions.

%% file: sections/method.tex
\section{Method}\label{sec:method}

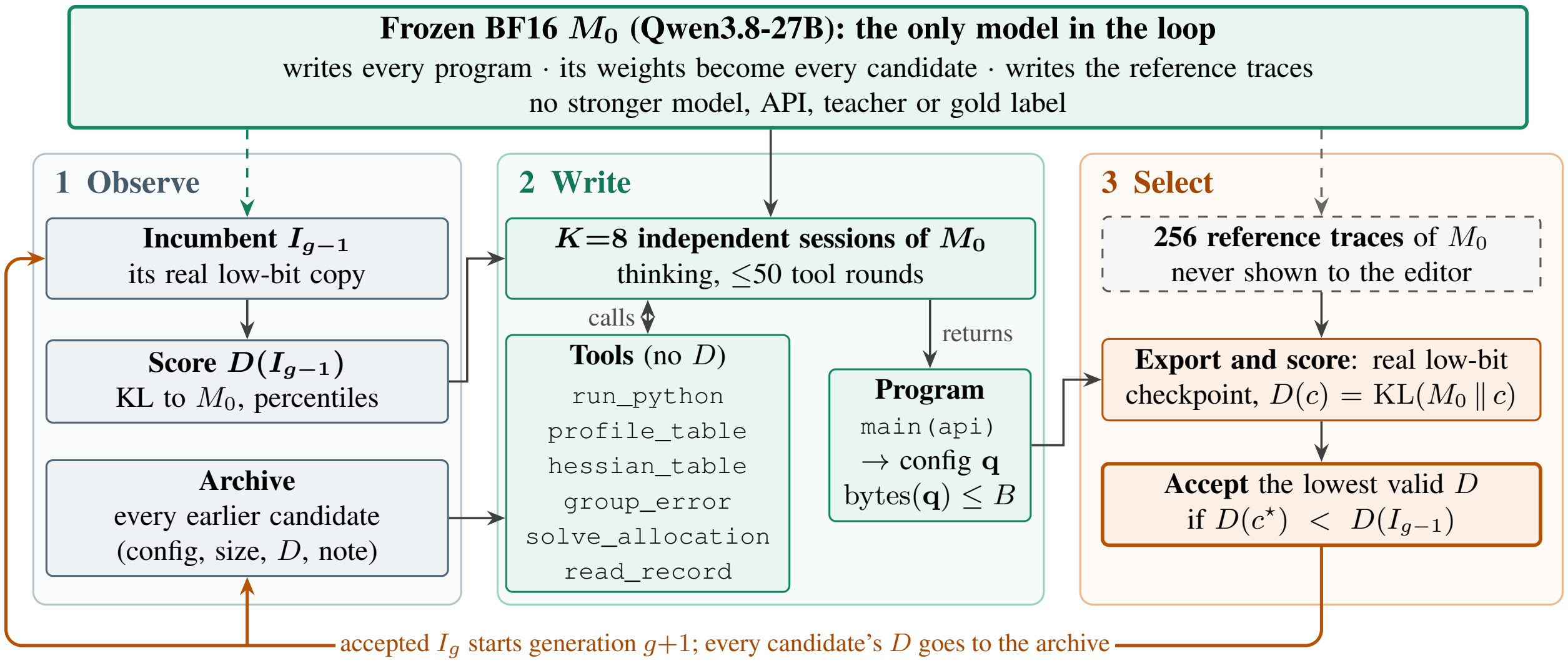
\begin{figure}[t]
  \centering
  \input{figures/overview}
  \caption{\textbf{One generation of \ours{}.} The editor reads the incumbent's KL to $\Mz$ and the archive of every
  earlier candidate (\emph{Observe}); eight independent sessions of the frozen $\Mz$ with self-measurement tools each
  return a program whose mixed-precision configuration fits $B$ (\emph{Write}); every candidate is exported as a real
  low-bit checkpoint, reads $\Mz$'s 256 fixed reference traces under teacher forcing, and the lowest KL below the
  incumbent's is accepted (\emph{Select}). The traces are never shown to the editor.}
  \label{fig:overview}
\end{figure}

In \ours{} the model quantizes itself. The frozen BF16 model $\Mz$ fills every role that needs intelligence
(\cref{fig:overview}): it writes the reference traces that define the objective, answers its own calibration prompts,
writes every candidate quantization, and supplies the original weights that every candidate quantizes. The object
that improves is $\Mz$'s quantized self within a fixed storage budget $B$, and the target is $\Mz$ itself: a candidate
is scored only by how closely it reproduces $\Mz$'s next-token distributions on 256 of $\Mz$'s own reasoning traces.
Fixed code builds, sizes and scores candidates; no other model, API, teacher or gold label takes part.

\subsection{Setting}\label{sec:method-setting}

\paragraph{Model and candidates.}
$\Mz$ is Qwen3.8-27B~\citep{qwen38}, a dense 64-layer decoder that uses gated full attention in every fourth layer and
Gated DeltaNet linear attention elsewhere. Its linear layers form $|\mathcal{K}|=368$ module groups: per layer the MLP
groups \texttt{gate\_up} and \texttt{down}, plus \texttt{qk}, \texttt{v} and \texttt{o} in full-attention layers or
four DeltaNet projections otherwise. A candidate is a weight-only mixed-precision configuration $\mathbf{q}$ that sets,
for every group $k$, the bit width $b_k\in\{2,3,4,8,16\}$ (16 keeps the group in BF16), the group size
$g_k\in\{32,64,128\}$ and the quantizer $a_k\in\{\textsc{rtn},\textsc{gptq}\}$~\citep{gptq} with asymmetric
per-group grids. Only the decoder's linear weights are quantized; the embedding and LM head stay in BF16. Groups that
the inference engine fuses into one matrix share a setting (\texttt{qk} with \texttt{v}, and the DeltaNet query--key--value
projection with its gate projection), so that every configuration is a deployable checkpoint. Every candidate
quantizes the \emph{original} weights, $Q(\Mz;\mathbf{q})$, so errors never compound across generations.

\paragraph{Budget.}
$\mathrm{bytes}(\mathbf{q})$ is the size of the packed checkpoint: for every quantized
$[d_{\mathrm{out}},d_{\mathrm{in}}]$ weight, the integer codes ($d_{\mathrm{out}}d_{\mathrm{in}}b/8$ bytes) and the
per-group scales and zero points, plus every other tensor of the model in BF16. The budget $B$ is the size of uniform
4-bit GPTQ with group size 128 on every linear group, $B=\budgetGiB$\,GiB. A candidate is valid only if
$\mathrm{bytes}(\mathbf{q})\le B$: the self-quantized model may never store more than uniform 4-bit weight
quantization. Each candidate is exported as a real low-bit checkpoint (integer codes with their scales, in GPTQ
format) and served by vLLM~\citep{vllm} with mixed-precision low-bit kernels; the score below is computed on exactly
the model that would be deployed.

\paragraph{Objective: KL to $\Mz$ on its own traces.}
Before the first generation, $\Mz$ writes one thinking-mode answer (temperature 0.6, top-$p$ 0.95, top-$k$ 20, at
most 16{,}384 tokens) to each of $n{=}256$ problems $X$ of a training pool, disjoint from the calibration texts and
from every evaluation set, and we cache $\Mz$'s top-20 next-token log-probabilities at every position. A candidate
$Q$ reads these fixed prompt--answer texts under teacher forcing, so nothing is sampled. For item $x$ with answer
positions $A_x$, let $T_t$ be $\Mz$'s top-20 tokens at position $t$, and let $p_t$ and $q_t$ be the distributions of
$\Mz$ and $Q$ on the partition $T_t\cup\{\text{rest}\}$. The score is
\begin{equation}\label{eq:kl}
  D(Q) = \frac{1}{|X|}\sum_{x\in X}\frac{1}{|A_x|}\sum_{t\in A_x}\ \sum_{v\in T_t\cup\{\text{rest}\}}
  p_t(v)\log\frac{p_t(v)}{q_t(v)},
\end{equation}
a lower bound on the full-vocabulary KL by the log-sum inequality~\citep{qda}. A reference token missing from $Q$'s
top-20 receives $Q$'s smallest reported log-probability. The task is
\begin{equation}\label{eq:objective}
  \min_{\mathbf{q}}\ D\big(Q(\Mz;\mathbf{q})\big)\quad\text{s.t.}\quad \mathrm{bytes}(\mathbf{q})\le B.
\end{equation}
$D$ needs no gold answer and no other model: $\Mz$ is both the reference and the judge of its own copy.

\paragraph{Self-measurement.}
Once, $\Mz$ answers 400 further pool problems greedily (at most 16{,}384 tokens), and each problem with its answer
forms a calibration text~\citep{selfcalib}, so calibration covers reasoning tokens and not only questions. The first
256 texts calibrate GPTQ and give per-layer Hessians and activation statistics. From them, code builds four fixed
tables of $\Mz$: a \emph{profile} table (whole-model KL increase when only one group is quantized to 2, 3 or 4 bits),
a \emph{Hessian} table (Hessian-weighted error $\operatorname{tr}(\Delta W H \Delta W^{\top})$ of each group,
quantizer and bit width), an \emph{activation} table (input outlier statistics) and a \emph{bytes} table.

\subsection{The self-quantization loop}\label{sec:method-loop}

Generation $g$ starts from the incumbent $I_{g-1}$. The loop starts from the BF16 model itself, $I_0=\Mz$, which has
$D=0$ but exceeds $B$; no quantized starting point is given, so $\Mz$ also chooses its first quantization.

\paragraph{Observe.}
Each session's context holds the model's layer and parameter summary, the incumbent's configuration with $D(I_{g-1})$
and the percentiles of its per-trace KL, the most sensitive groups of the profile table, activation outlier flags,
and an archive with one row per earlier candidate (configuration, size, $D$ and the session's own note). Tools open any
earlier candidate with its program and feedback, including the groups with the largest Hessian-weighted error. The
editor never sees the reference traces, only scores computed on them.

\paragraph{Write.}
$K{=}8$ independent thinking-mode sessions of $\Mz$ run in parallel; a session does not see the others' candidates of
the same generation. Each may call tools for up to 50 rounds, with a 131{,}072-token context, and then returns a
program \texttt{def main(api)} that outputs a configuration. \Cref{tab:tools} lists the tools. A persistent Python
sandbox runs exploratory code with the same \texttt{api} as the final program: it reads the four tables, sizes a
configuration, computes the Hessian-weighted error of any configuration per group (\texttt{group\_error}), and solves
a multiple-choice knapsack over per-group (bytes, damage) trade-offs with inputs the editor chooses
(\texttt{solve\_allocation}). The editor decides what to optimize, and code does the arithmetic. No tool evaluates $D$.

A configuration has a default setting (bits, group size, quantizer), an ordered list of rules that override it for
layer ranges or group types (for example \texttt{60-63.down}), and per-group overrides applied last, so a program can
express anything from a uniform setting to a different choice for each of the 368 groups. Programs run in an isolated
subprocess without network access, with 16\,GB of memory and 1{,}800\,s of wall-clock time; tool calls are answered
by the trusted parent process, which validates the configuration, computes its size and rejects it if it exceeds $B$.
Programs never see the model weights or the reference traces. A failed, invalid or over-budget program yields no
candidate.

\begin{table}[t]
\centering
\small
\caption{\textbf{The editor's tools.} The tables are built once from $\Mz$ before generation 1; no tool evaluates the
selection score $D$.}
\label{tab:tools}
\begin{tabular*}{\textwidth}{@{\extracolsep{\fill}}p{0.33\textwidth}p{0.62\textwidth}@{}}
\toprule
Tool & Returns \\
\midrule
\texttt{model\_info()} & layer types, groups per layer type, parameters per group, budget $B$, action space and the
incumbent's configuration \\
\texttt{profile\_table()}, \texttt{profile\_summary()} & whole-model KL increase when only one group is quantized to
2, 3 or 4 bits; the most and least sensitive groups \\
\texttt{hessian\_table()} & Hessian-weighted error of quantizing each group with each quantizer and bit width \\
\texttt{activation\_table()} & per module input: RMS statistics, outlier fraction and an outlier flag \\
\texttt{bytes\_table()}, \texttt{bytes(config)} & packed size of every group at every setting; validity, size,
average bits and budget check of a configuration \\
\texttt{group\_error(config, top)} & per-group Hessian-weighted error of a configuration and its largest contributors \\
\texttt{solve\_allocation(\dots)} & the configuration minimizing a weighted sum of per-group damage under a byte
budget, for weights, bit widths, group sizes, quantizer and fixed groups chosen by the editor \\
\texttt{run\_python(code)} & runs exploratory code with this API in a persistent sandbox \\
\texttt{read\_record(id)}, \texttt{search\_history()} & the full program and feedback of an earlier candidate; one
line per earlier candidate \\
\bottomrule
\end{tabular*}
\end{table}

\subsection{Selection by KL}\label{sec:method-gate}

Every valid candidate $c$ is exported, served and scored once by \cref{eq:kl}. The candidate with the lowest score,
$c^\star=\arg\min_c D(c)$, becomes $I_g$ if $D(c^\star)<D(I_{g-1})$; otherwise $I_g=I_{g-1}$. In generation 1, where
$I_0=\Mz$ is over budget, the lowest-scoring valid candidate is accepted. Every candidate's score, percentiles and top
error groups are stored, and later generations read them, accepted or not. The loop runs a fixed number of
generations without early stopping (\cref{alg:selfquant} in \cref{sec:app-alg}). Task accuracy is measured outside the loop, on
benchmarks the loop never sees (\cref{sec:exp}).

\subsection{Design choices}\label{sec:method-design}

\paragraph{Why KL to $\Mz$ on its own traces.}
The selection signal must come from the model itself and must not reward anything beyond $\Mz$'s own competence. The
KL to $\Mz$ meets both conditions: its minimum, $D=0$, is $\Mz$ itself, so the objective cannot be satisfied by
imitating a stronger model or by fitting a benchmark, and it needs neither gold answers nor a judge. Computing it on
$\Mz$'s own thinking-mode answers, rather than on generic text, measures the quantized model on the long reasoning
sequences it will produce when deployed, where quantization damage accumulates \citep{quanthurtsreasoning,lotfi}.
Teacher forcing on fixed traces makes $D$ deterministic for a given checkpoint: two candidates are compared on
exactly the same tokens, and a score never depends on sampling luck.

\paragraph{Why start from BF16.}
Starting from a uniform 4-bit configuration would hand the loop a strong human-designed solution and blur what the
model contributes. Starting from $I_0=\Mz$, the first quantization is written by the model itself, and every later
incumbent descends from its own decisions. Uniform GPTQ, which defines $B$, is used only to set the budget and as a
reference row in the results; it is never shown to the editor as a configuration.

\paragraph{Why a frozen editor, independent sessions and an archive.}
The editor is the original BF16 model in every generation, so its competence is fixed and every gain comes from what
it learns about its own weights through tools and feedback, not from a changed model. Independent sessions explore
different allocations in parallel without converging on one session's idea, and the archive turns every scored
candidate, accepted or not, into evidence for the next generation: each session sees which configurations were tried,
how large they were and how far each was from $\Mz$.

%% file: figures/overview.tex
\resizebox{\textwidth}{!}{%
\begin{tikzpicture}[
  font=\footnotesize,
  >={Stealth[length=2.0mm,width=1.6mm]},
  box/.style={draw=#1!85!black, fill=#1!10, rounded corners=2pt, align=center, inner sep=3pt, line width=0.6pt},
  flow/.style={->, line width=0.7pt, draw=black!75},
  lab/.style={font=\scriptsize, text=black!70, fill=white, inner sep=1pt},
  block/.style={draw=#1!45, fill=#1!4, rounded corners=4pt, line width=0.6pt},
  btitle/.style={font=\small\bfseries, text=#1!75!black},
]
\definecolor{dteal}{HTML}{1B9E77}
\definecolor{dorange}{HTML}{D95F02}
\definecolor{dgrey}{HTML}{666666}
\definecolor{dblue}{HTML}{5E7A8C}
\def\tq#1{{\ttfamily\scriptsize #1}}

\begin{scope}[on background layer]
  \draw[block=dblue]   (0.00,1.15) rectangle (4.70,6.10);
  \draw[block=dteal]   (5.10,1.15) rectangle (11.10,6.10);
  \draw[block=dorange] (11.50,1.15) rectangle (16.80,6.10);
\end{scope}
\node[btitle=dblue,   anchor=north west] at (0.10,6.05)  {1\enspace Observe};
\node[btitle=dteal,   anchor=north west] at (5.20,6.05)  {2\enspace Write};
\node[btitle=dorange, anchor=north west] at (11.60,6.05) {3\enspace Select};

\node[box=dteal, line width=1.1pt, text width=15.8cm] (m0) at (8.40,7.05)
  {\textbf{\small Frozen BF16 $\boldsymbol{M_0}$ (Qwen3.8-27B): the only model in the loop}\\[1pt]
   writes every program $\cdot$ its weights become every candidate $\cdot$ writes the reference traces\\
   no stronger model, API, teacher or gold label};

\node[box=dblue, text width=4.2cm] (inc) at (2.35,4.95)
  {\textbf{Incumbent $\boldsymbol{I_{g-1}}$}\\ its real low-bit copy};
\node[box=dblue, text width=4.2cm] (fb) at (2.35,3.60)
  {\textbf{Score $\boldsymbol{D(I_{g-1})}$}\\ KL to $M_0$, percentiles};
\node[box=dblue, text width=4.2cm] (arch) at (2.35,2.10)
  {\textbf{Archive}\\ every earlier candidate\\ (config, size, $D$, note)};
\draw[flow] (inc) -- (fb);

\node[box=dteal, text width=5.6cm] (sess) at (8.10,4.95)
  {\textbf{$\boldsymbol{K{=}8}$ independent sessions of $\boldsymbol{M_0}$}\\ thinking, $\le$50 tool rounds};
\node[box=dteal, text width=2.9cm] (tools) at (6.75,2.70)
  {\textbf{Tools} (no $D$)\\[1pt]
   \tq{run\_python}\\ \tq{profile\_table}\\ \tq{hessian\_table}\\ \tq{group\_error}\\ \tq{solve\_allocation}\\
   \tq{read\_record}};
\node[box=dteal, text width=2.0cm] (prog) at (9.85,2.90)
  {\textbf{Program}\\ \tq{main(api)}\\ $\to$ config $\mathbf{q}$\\ $\mathrm{bytes}(\mathbf{q})\le B$};
\draw[<->, line width=0.7pt, draw=black!75] (tools.north |- sess.south) -- (tools.north);
\draw[flow] (prog.north |- sess.south) -- (prog.north);
\node[font=\scriptsize, text=black!70, anchor=east] at ($(tools.north |- sess.south)!0.5!(tools.north)$) {calls};
\node[font=\scriptsize, text=black!70, anchor=west] at ($(prog.north |- sess.south)!0.5!(prog.north)$) {returns};

\draw[flow] (fb.east) -- ++(0.20,0) |- (sess.west);
\draw[flow] (arch.east) -- (arch.east -| tools.west);

\node[box=dgrey, dashed, fill=dgrey!6, text width=4.6cm] (refs) at (14.15,5.00)
  {\textbf{256 reference traces} of $M_0$\\ never shown to the editor};
\node[box=dorange, text width=4.6cm] (score) at (14.15,3.62)
  {\textbf{Export and score}: real low-bit\\ checkpoint, $D(c)=\mathrm{KL}(M_0\,\|\,c)$};
\node[box=dorange, line width=1.1pt, text width=4.6cm] (acc) at (14.15,2.25)
  {\textbf{Accept} the lowest valid $D$\\ if $D(c^\star)<D(I_{g-1})$};
\draw[flow] (refs) -- (score);
\draw[flow] (score) -- (acc);
\draw[flow] (prog.east) -- ++(0.35,0) |- (score.west);

\draw[flow] (m0.south -| sess.north) -- (sess.north);
\draw[->, line width=0.7pt, dashed, draw=dteal!80!black] (m0.south -| inc.north) -- (inc.north);
\draw[->, line width=0.7pt, dashed, draw=dgrey] (m0.south -| refs.north) -- (refs.north);

\draw[flow, dorange!85!black, line width=1.0pt, rounded corners=5pt]
  (acc.south) -- (14.15,0.70) -- (-0.30,0.70) |- (inc.west);
\draw[flow, dorange!85!black, line width=1.0pt] (arch.south |- 0,0.70) -- (arch.south);
\node[font=\scriptsize, text=dorange!75!black, fill=white, inner sep=1.5pt] at (7.60,0.70)
  {accepted $I_g$ starts generation $g{+}1$; every candidate's $D$ goes to the archive};
\end{tikzpicture}%
}

%% file: sections/experiments.tex
\section{Experimental Setup}\label{sec:exp}

\paragraph{Model, budget and loop.}
$\Mz$ is Qwen3.8-27B~\citep{qwen38} in BF16 (\cref{sec:method-setting}). The model quantizes itself with mixed
precision over $\{2,3,4,8,16\}$ bits per module group, under the storage budget of uniform 4-bit GPTQ~\citep{gptq}
with group size 128 ($B=\budgetGiB$\,GiB, embedding and LM head in BF16). The 256 KL reference traces and the 400
calibration texts come from the DAPO-Math-17K~\citep{dapo} training pool, decontaminated against the evaluation sets,
and are answered by $\Mz$ itself, so no gold answer enters the loop. The loop starts from the BF16 model and runs five
generations; each generation runs $K{=}8$ editor sessions of the frozen $\Mz$ and scores every valid candidate by
\cref{eq:kl} on the same 256 traces. Every candidate is a real packed low-bit checkpoint served by vLLM~\citep{vllm} on
one GPU; the run uses eight NVIDIA H20 GPUs, all of which serve the editor while it writes and the candidates while
they are scored.

\paragraph{Implementation.}
A candidate is exported as a GPTQ-format checkpoint: integer codes packed into 32-bit words, BF16 scales and packed
zero points per group, and BF16 tensors for groups kept at 16 bits. vLLM~0.24~\citep{vllm} serves these checkpoints
with mixed-precision low-bit kernels, so the score $D$ and the reported accuracies are those of the deployable
low-bit model. GPTQ and the Hessian table use the 256 self-generated calibration texts of up to 18{,}432 tokens each,
so the Hessians cover long reasoning answers and not only prompts. The tables of \cref{tab:tools} are built once before
generation 1 and reused by every generation; every candidate is quantized from $\Mz$'s original weights through a
shared layer cache, so two candidates that choose the same setting for a group share the same quantized weights.

\paragraph{Baselines.}
We compare with three uniform one-shot quantizers at 4 bits and group size 128, built from the same self-generated
calibration texts: RTN, GPTQ~\citep{gptq}, whose packed size defines $B$, and AWQ-lite, our implementation of
AWQ~\citep{awq} that applies activation-aware channel scaling on the RTN grid. Like every \ours{} candidate, each baseline is
exported as a real packed low-bit checkpoint and served with the same low-bit kernels. The BF16 model $\Mz$ is the ceiling of any quantization of itself.

\paragraph{Evaluation.}
We evaluate on AIME24 and AIME25 (30 problems each), which the loop never sees. For each problem a model samples 16
answers in Qwen3 thinking mode~\citep{qwen3} (temperature 0.6, top-$p$ 0.95, top-$k$ 20, at most 32{,}768 output
tokens, thinking included), and we report avg@16, the mean accuracy over the 16 answers, graded by deterministic
math-answer equivalence. Every quantized model is evaluated from its real packed low-bit checkpoint; the model
evaluated for \ours{} is the incumbent after five generations, which takes $\finalGiB$\,GiB ($\le B$) against
$\sizeMz$\,GiB for $\Mz$.

%% file: sections/results.tex
\section{Results}\label{sec:results}

\paragraph{KL to $\Mz$ across generations.}
\Cref{tab:kl} and \cref{fig:kl} give, for each generation, the KL of the incumbent the generation starts from, the
lowest KL among its valid candidates, and the KL of the incumbent it ends with. Generation 1 starts from the BF16 model
($D=0$, over budget), so its best in-budget candidate, with $D=\klIncA$, is the model's first quantization of itself.
In every later generation the best candidate is below the incumbent and is accepted, so the KL falls in every
generation, to $\klIncE$ after generation 5, $\klDropGen$\% below generation 1. The largest step comes in generation 2
($-17.4\%$), when the sessions first read the scores of earlier candidates. Uniform 4-bit GPTQ with group size 128,
whose size defines $B$ and which is never given to the loop, has $D=\klQz$: the model's own quantization passes it in
generation 3 and ends $\klBelowQz$\% below it after generation 5.

\begin{figure}[t]
\centering
\begin{minipage}[c]{0.485\textwidth}
\centering
\begin{tikzpicture}
\begin{axis}[
  width=\linewidth, height=4.6cm,
  xlabel={Generation}, ylabel={KL to $M_0$ ($\times 10^{-3}$)},
  xmin=0.6, xmax=5.4, ymin=2.8, ymax=4.05,
  xtick={1,2,3,4,5},
  ytick={2.8,3.0,3.2,3.4,3.6,3.8,4.0},
  yticklabel style={/pgf/number format/fixed,/pgf/number format/precision=1,/pgf/number format/fixed zerofill},
  tick label style={font=\scriptsize}, label style={font=\small},
  grid=major, grid style={draw=black!10},
  every axis plot/.append style={line width=1.0pt},
]
\addplot[color=black!45, dashed, domain=0.6:5.4, samples=2, line width=0.8pt] {3.263};
\node[anchor=south east, font=\scriptsize, text=black!60] at (axis cs:5.35,3.27) {uniform GPTQ W4};
\addplot[color={rgb,255:red,27;green,158;blue,119}, mark=*, mark size=1.8pt] coordinates {
  (1,3.956) (2,3.269) (3,3.061) (4,2.921) (5,2.908)};
\end{axis}
\end{tikzpicture}
\captionof{figure}{\textbf{Best candidate KL per generation} on the 256 fixed reference traces (lower is better);
each was accepted as the new incumbent. Dashed: uniform 4-bit GPTQ.}
\label{fig:kl}
\end{minipage}\hfill
\begin{minipage}[c]{0.485\textwidth}
\centering
\small
\captionof{table}{\textbf{KL to $\Mz$ per generation} (\cref{eq:kl}). \emph{In}/\emph{out}: incumbent at the start/end
of the generation; \emph{valid}: candidates within $B$ and scored, out of $K{=}8$.}
\label{tab:kl}
\begin{tabular*}{\linewidth}{@{\extracolsep{\fill}}ccccc@{}}
\toprule
Gen. & In & Best candidate & Out & Valid \\
\midrule
1 & 0 (BF16) & \klBestA & \klIncA & \validA/8 \\
2 & \klIncA & \klBestB & \klIncB & \validB/8 \\
3 & \klIncB & \klBestC & \klIncC & \validC/8 \\
4 & \klIncC & \klBestD & \klIncD & \validD/8 \\
5 & \klIncD & \klBestE & \textbf{\klIncE} & \validE/8 \\
\midrule
\multicolumn{5}{@{}l}{Uniform GPTQ W4, $g$128 (not given to the loop): \klQz} \\
\bottomrule
\end{tabular*}
\end{minipage}
\end{figure}

\paragraph{Accuracy on AIME.}
\Cref{tab:aime} reports avg@16 on AIME24 and AIME25. After five generations, the self-quantized model stores its linear weights at
$\finalBits$ bits on average, in $\finalGiB$\,GiB against $\sizeMz$\,GiB for BF16, and reaches $\aimeSqAvg$\% on
average, $\aimeRetain$\% of the BF16 model's $\aimeMzAvg$\%. With fewer bits per linear weight on average
than the uniform 4-bit baselines, it is the most accurate 4-bit-budget model on both AIME24 and AIME25, ahead of GPTQ,
AWQ-lite and RTN; uniform RTN, in contrast, loses 10.21 points on AIME25. The
loop never sees AIME problems or gold answers: its only signal is the KL to $\Mz$ on $\Mz$'s own traces.

\paragraph{The self-referential signal transfers to held-out accuracy.}
The two measurements are independent: $D$ is computed by teacher forcing on 256 DAPO-Math traces written by $\Mz$,
while AIME accuracy is measured by free generation of up to 32{,}768 tokens on problems from a different source that
the loop never sees. They agree. The self-quantized model has a lower KL to $\Mz$ than uniform GPTQ ($\klIncE$ against
$\klQz$) and a higher AIME accuracy on both sets. Lowering the divergence from its own distribution, the only target the model is given, thus moves
the quantized model toward the BF16 model's competence on unseen problems.

\paragraph{Where the bits go.}
The final configuration averages $\finalBits$ bits per linear weight, below the 4 bits of every uniform baseline,
while its packed size stays within the same budget $B$. Because the budget counts every byte of the packed checkpoint,
including per-group scales and zero points, a self-written configuration can trade code bits against what a uniform
format cannot express: different bit widths over $\{2,3,4,8,16\}$ for different groups, different group sizes, and
the choice between RTN and GPTQ for each group. A uniform 4-bit format spends every byte the same way; the
self-written configuration spends them where the model's own measurements say they matter.

\begin{table}[t]
\centering
\small
\caption{\textbf{AIME accuracy} (avg@16, \%; 30 problems per set, 16 answers each). \emph{Bits}: parameter-weighted
average bits of the linear weights. The 4-bit baselines are uniform with group size 128; \ours{} uses mixed precision within the same storage budget. Every quantized model is evaluated
from its real packed low-bit checkpoint.}
\label{tab:aime}
\begin{tabular*}{\textwidth}{@{\extracolsep{\fill}}lcccc@{}}
\toprule
Model & Bits & AIME24 & AIME25 & Avg. \\
\midrule
BF16 $\Mz$ (ceiling) & 16 & \aimeMzA & \aimeMzB & \aimeMzAvg \\
\midrule
RTN W4 & 4 & \aimeRtnA & \aimeRtnB & \aimeRtnAvg \\
GPTQ W4 & 4 & \aimeGptqA & \aimeGptqB & \aimeGptqAvg \\
AWQ-lite W4 & 4 & \aimeAwqA & \aimeAwqB & \aimeAwqAvg \\
\ours{} (generation 5) & \finalBits & \textbf{\aimeSqA} & \textbf{\aimeSqB} & \textbf{\aimeSqAvg} \\
\bottomrule
\end{tabular*}
\end{table}

%% file: sections/discussion.tex
\section{Discussion}\label{sec:discussion}

\paragraph{Who makes the decisions.}
The tools of \cref{tab:tools} measure and do arithmetic: they size configurations, report errors and solve a knapsack
for the weights and constraints they are given. What to measure, which signal to trust, which groups to protect, how
to weigh the profile table against the Hessian table, which bit widths, group sizes and quantizers to allow, and how
to react to the scores in the archive are all decided in the programs that $\Mz$ writes. No tool returns the
selection score, and no configuration enters the loop except through a program written by $\Mz$.

%% file: sections/conclusion.tex
\section{Conclusion}\label{sec:conclusion}

We asked whether a model can quantize itself with no stronger model, API, teacher or gold label in the loop. In
\ours{}, the frozen BF16 model starts from its own weights, measures them with tools, writes mixed-precision
configurations for its real low-bit copy, and judges each copy by its KL divergence on its own reasoning traces. On
Qwen3.8-27B, within the storage of uniform 4-bit weights, five generations end $\klBelowQz$\% below uniform GPTQ in
KL, and the self-quantized model reaches $\aimeSqAvg$\% avg@16 on AIME24 and AIME25, $\aimeRetain$\% of the BF16 model
and above uniform 4-bit RTN, GPTQ and AWQ-lite. A model can improve its deployable self using only itself.

%% file: sections/appendix.tex
\section{Algorithm and Settings}\label{sec:app-hparams}\label{sec:app-alg}

\Cref{alg:selfquant} gives the loop of \cref{sec:method-loop,sec:method-gate}, and \cref{tab:hparams} lists the
settings of the reported run.

\begin{algorithm}[!ht]
\caption{\ours{}: the self-quantization loop}\label{alg:selfquant}
\small
\begin{algorithmic}[1]
\Require frozen BF16 model $\Mz$; budget $B$; generations $G$; sessions $K$
\State $\Mz$ writes 256 reference traces $X$ (top-20 log-probabilities cached) and 400 calibration answers
\State build the profile, Hessian, activation and bytes tables of $\Mz$; archive $\mathcal{A}\gets\emptyset$
\State $I_0\gets\Mz$, $D(I_0)\gets 0$ \Comment{start from BF16; $I_0$ exceeds $B$}
\For{$g = 1, \dots, G$}
  \For{sessions $k = 1, \dots, K$ \textbf{in parallel}}
    \State $\Mz$ reads $D(I_{g-1})$ and $\mathcal{A}$, calls tools ($\le 50$ rounds), returns a program $\pi_k$
    \State $\mathbf{q}_k\gets\textsc{Sandbox}(\pi_k)$; \textbf{if} valid and $\mathrm{bytes}(\mathbf{q}_k)\le B$
    \textbf{then} export $Q(\Mz;\mathbf{q}_k)$ as a low-bit checkpoint and score $D(\mathbf{q}_k)$ (\cref{eq:kl})
  \EndFor
  \State $c^\star\gets\arg\min_k D(\mathbf{q}_k)$
  \If{$g=1$ \textbf{or} $D(c^\star)<D(I_{g-1})$}
    \State $I_g\gets c^\star$
  \Else
    \State $I_g\gets I_{g-1}$
  \EndIf
  \State $\mathcal{A}\gets\mathcal{A}\cup\{$every candidate's configuration, size, $D$, percentiles, note$\}$
\EndFor
\State \Return $I_G$
\end{algorithmic}
\end{algorithm}

\begin{table}[!ht]
\centering
\footnotesize
\renewcommand{\arraystretch}{0.95}
\caption{Settings of the reported run.}
\label{tab:hparams}
\begin{tabular}{@{}p{0.30\linewidth}p{0.64\linewidth}@{}}
\toprule
Setting & Value \\
\midrule
\multicolumn{2}{@{}l}{\emph{Model and editor}} \\
Model $\Mz$ & Qwen3.8-27B, BF16 \\
Editor & frozen $\Mz$ in every generation; thinking mode; temperature 0.6, top-$p$ 0.95, top-$k$ 20 \\
Editor limits & 131{,}072-token context; $\le$50 tool rounds of $\le$16{,}384 tokens each; final turn
(thinking and program) $\le$32{,}768 tokens \\
Loop & start from BF16; $G=5$ generations, no early stopping; $K=8$ sessions $\times$ 1 candidate \\
Program sandbox & no network; \texttt{run\_python}: 300\,s and 8\,GB per call \\
\midrule
\multicolumn{2}{@{}l}{\emph{Action space and budget}} \\
Quantized weights & the decoder's linear weights; embedding and LM head in BF16 \\
Bits / group sizes & $\{2,3,4,8,16\}$ (16 = BF16) / $\{32,64,128\}$, per module group \\
Quantizers & RTN, GPTQ (damping 0.01), asymmetric per-group grids \\
Budget $B$ & packed size of uniform 4-bit GPTQ, $g$128: \budgetGiB\,GiB \\
Checkpoint & GPTQ-format packed integer codes with scales and zero points; fused modules share a setting \\
\midrule
\multicolumn{2}{@{}l}{\emph{Data and objective}} \\
Pool & DAPO-Math-17K, decontaminated against AIME24 and AIME25 \\
KL reference traces & 256 problems, one $\Mz$ answer each (thinking; temperature 0.6, top-$p$ 0.95, top-$k$ 20;
$\le$16{,}384 tokens); top-20 log-probabilities \\
Calibration texts & 400 further problems with $\Mz$'s greedy answer ($\le$16{,}384 tokens); 256 calibrate GPTQ \\
Profile table & one group at a time at 2/3/4 bits, $g$128; 64 texts of $\le$512 tokens \\
\midrule
\multicolumn{2}{@{}l}{\emph{Evaluation}} \\
Benchmarks & AIME24, AIME25 (30 problems each) \\
Decoding & thinking; temperature 0.6, top-$p$ 0.95, top-$k$ 20; $\le$32{,}768 output tokens; 16 samples
(avg@16) \\
\midrule
\multicolumn{2}{@{}l}{\emph{Hardware and software}} \\
GPUs & 8 NVIDIA H20; the editor runs tensor parallel on all 8 and sleeps while the candidates, one per GPU, are scored \\
Software & vLLM 0.24.0 with mixed-precision low-bit (Humming) kernels \\
\bottomrule
\end{tabular}
\end{table}